\documentclass[letterpaper, 10 pt, conference]{ieeeconf}  % Comment this line out if you need a4paper

\IEEEoverridecommandlockouts                              % This command is only needed if 
\usepackage{graphics} % for pdf, bitmapped graphics files
\usepackage{epsfig} % for postscript graphics files
\usepackage{times} % assumes new font selection scheme installed
\usepackage{amsmath} % assumes amsmath package installed
\usepackage{amssymb}  % assumes amsmath package installed

\usepackage{subcaption}
\usepackage{wrapfig}
\usepackage{algorithm}
\usepackage{algpseudocode}
\usepackage{amsmath}
\usepackage{amssymb}
\usepackage{xcolor}

\usepackage{graphicx}
\graphicspath{{./images}}

\newcommand{\mdp}[1]{\mathcal{#1}}

\title{\LARGE \bf
SLIM: Skill Learning with Multiple Critics
}

\author{
David Emukpere$^{1}$, Bingbing Wu$^{1}$, Julien Perez$^{2\dagger}$, Jean-Michel Renders$^{1}$% <-this % stops a space
\thanks{$^{1}$NAVER LABS Europe, 6 chemin de Maupertuis, Meylan, 38240, France. email: {\tt\small firstname.lastname@naverlabs.com}}
\thanks{$^{2}$EPITA Research  Laboratory (LRE), FR-94276 Le Kremlin-Bic\^etre, France.  email: {\tt\small firstname.lastname@epita.fr}}
\thanks{$^{\dagger}$Work done at Naver Labs Europe}
}

\begin{document}

\maketitle
\thispagestyle{empty}
\pagestyle{empty}

\begin{abstract}

Self-supervised skill learning aims to acquire useful behaviors that leverage the underlying dynamics of the environment. 
Latent variable models, based on mutual information maximization, have been successful in this task but still struggle in the context of robotic manipulation. 
As it requires impacting a possibly large set of degrees of freedom composing the environment, mutual information maximization fails alone in producing useful and safe manipulation behaviors. Furthermore, tackling this by augmenting skill discovery rewards with additional rewards through a naive combination might fail to produce desired behaviors. To address this limitation, we introduce SLIM, a multi-critic learning approach for skill discovery with a particular focus on robotic manipulation. Our main insight is that utilizing multiple critics in an actor-critic framework to gracefully combine multiple reward functions leads to a significant improvement in latent-variable skill discovery for robotic manipulation while overcoming possible interference occurring among rewards which hinders convergence to useful skills. Furthermore, in the context of tabletop manipulation, we demonstrate the applicability of our novel skill discovery approach to acquire safe and efficient motor primitives in a hierarchical reinforcement learning fashion and leverage them through planning, significantly surpassing baseline approaches for skill discovery.

\end{abstract}

\section{Introduction}

%1st paragraph: Why is the problem important?

Self-supervised methods for skill discovery have been extensively developed in recent years as they enable robots to acquire reusable and transferable knowledge.
This flexibility is crucial in dynamic and unstructured environments where robots encounter variations, uncertainties, and unforeseen events.
Instead of engineering explicit rules and behaviors for each individual task, robots can learn from data and experiences, making the learning process scalable, thus improving the efficiency and versatility of robotic systems.

%2nd paragraph: Why is the problem difficult?

One popular approach to skill discovery utilizes the so-called mutual information maximization objective~\cite{Barber2003TheIA} to derive intrinsic rewards~\cite{Gregor2016VariationalIC}. 
Commonly, this involves training a latent-variable conditioned policy with reinforcement learning which maximizes the mutual information between the latent variable i.e. \textit{skill}, given as input to the agent, and the agent's state~\cite{Gregor2016VariationalIC, Sharma2019DynamicsAwareUD}.
While this formulation has been shown to enable an embodied agent to discover behaviors with respect to changing its own state, as in locomotion~\cite{Eysenbach2018DiversityIA}, it struggles in situations where we desire to discover skills that affect degrees of freedom composing the state space outside the agent's own state.
For example, impacting object states, as is the case in robotic manipulation, would require extensive exploration to discover interaction skills.

\begin{figure}[t]
    \centering
    \begin{subfigure}{0.36\textwidth}
    \centering
    \includegraphics[height=4.0cm, width=1.0\linewidth]{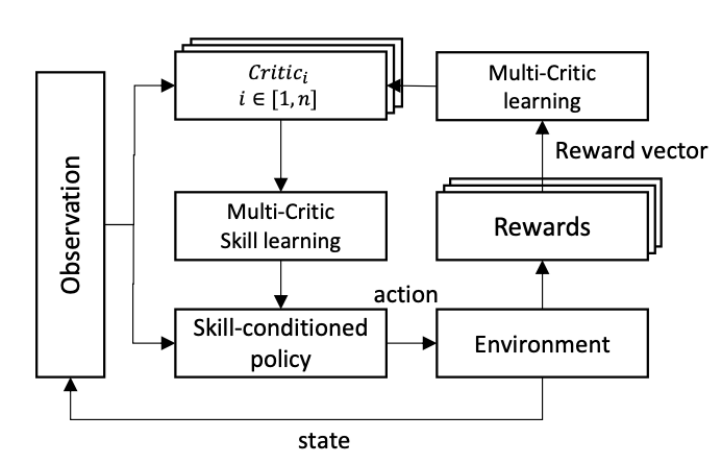}
    \caption{}
    \label{fig:slim_schematic}
    \end{subfigure}
    \begin{subfigure}{0.18\textwidth}
    \centering
    \includegraphics[height=3.0cm, width=1.0\linewidth]{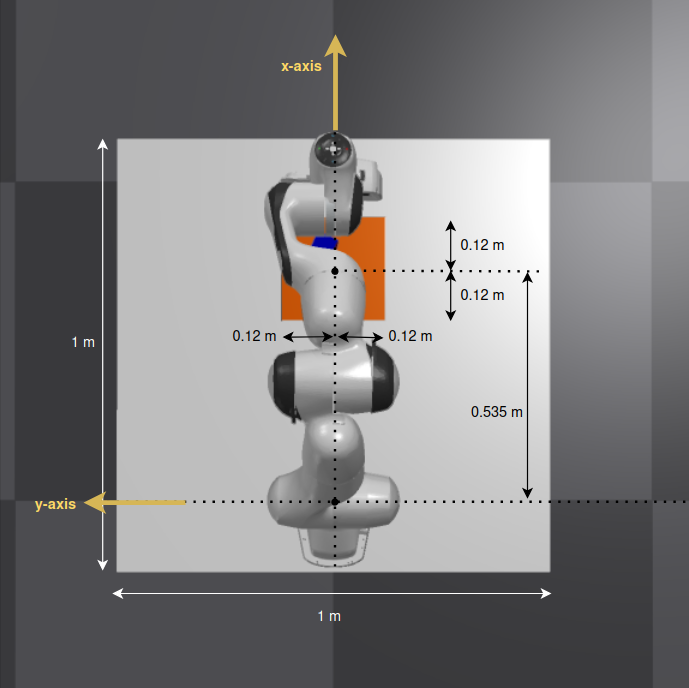}
    \caption{}
    \label{fig:slim_sim_topview}
    \end{subfigure}
    \begin{subfigure}{0.18\textwidth}
    \centering
    \includegraphics[height=3.0cm, width=1.0\linewidth]{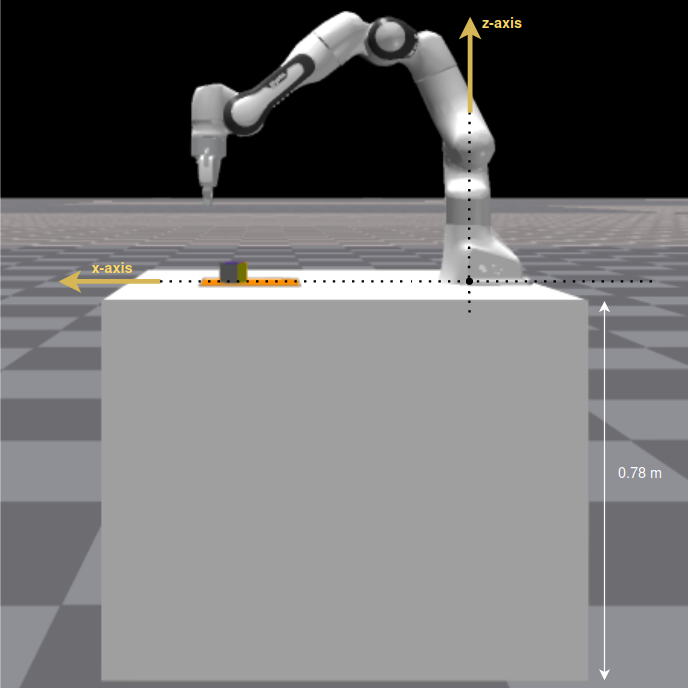}
    \caption{}
    \label{fig:slim_sim_sideview}
    \end{subfigure}
    \caption{\textbf{Skill Learning wIth Multiple critics}. Our approach enables the effective combination of multiple objectives for self-supervised skill discovery in robotic manipulation. We learn dedicated critics per intrinsic reward function which is used during policy improvement by taking a weighted combination of their normalized advantages.
    (a) Schematic diagram (b) Simulation top view (c) Simulation side view. }
    \label{fig:slim}
\end{figure}

A simple way to tackle this challenging case might be to augment intrinsic rewards with additional components that encourage such interactions within the environment. 
For example, one can introduce a reaching bonus to reward the end-effector of the considered manipulator to get close to the objects composing the considered scene. 
Furthermore, ensuring the safety of skill discovery is another critical problem recently introduced in ~\cite{kim2023safeskill}. 
Adding this also defines an extra reward component that needs to be carefully combined with other reward terms. 
In this work, we show that a naive implementation of combining these multiple rewards to obtain meaningful and safe interaction skills typically doesn't work or, in the best case, would require laborious tuning to derive a weighted combination that elicits the desired behavior. 
To solve this, we introduce a novel multi-critic~\cite{Mysore2022MultiCriticAL} approach to self-supervised skill discovery that is simple to implement and requires little to no effort to find the right combination of different rewards for safe and effective robotic manipulation skill discovery.
 
We demonstrate the applicability of our approach for acquiring safe and effective motor primitives in a hierarchical reinforcement learning (HRL) fashion.
Then, we leverage them for rearrangement and object trajectory following tasks through planning, surpassing the state-of-the-art baseline approaches for skill discovery. \\ ~ \\

In summary, our main contributions are:
\begin{itemize}
    \item We introduce SLIM, a robust multi-critic approach to latent variable skill discovery which enables us to train skill-conditioned policies with useful, diverse, and safe behaviors.
    \item We perform extensive ablation tests that illustrate the benefits of our approach for skill discovery.
    \item We evaluate SLIM against the main state-of-the-art approaches of skill discovery in challenging robotic manipulation scenarios.
    \item We demonstrate the benefit of SLIM for training HRL-based motor primitives used for object-centric trajectory tracking.
\end{itemize}

\section{Related Work}
\label{sec:related}

\subsection{Skill Discovery}

Numerous skill discovery methods~\cite{Sharma2019DynamicsAwareUD, Eysenbach2018DiversityIA, Lee2019SMM, Campos2020EDL, Lee2020LearningTC, Liu2021APT, Zhu2021BottomUpSD, laskin_unsupervised_2022} and benchmarks~\cite{Laskin2021URLBUR} for robotics have been actively studied in recent years due to the perceived fruitfulness of unsupervised pretraining for efficient adaptation to new tasks. In general,  while most of these methods have produced impressive results in various robotics domains such as locomotion, navigation and simple manipulation settings, for example with narrow initial state distributions and fixed end-effector orientation, they typically struggle in more challenging manipulation environments. One reason for this is the extended exploration due to wider initial state distributions and larger action space (position \textit{and} orientation) needed to learn consistent interaction with objects. Without these interactions, obtaining intrinsic rewards related to diversity in object manipulation becomes infeasible. In addition, mutual information maximization done without any prior information between the proprioception and exteroception parts of the state definition leads to local minima which leads to agents mostly moving around their embodiment with little environmental impact~\cite{Park2022LSD, Park2023CSD}. 

There exists a few interesting approaches to mitigating this problem with skill discovery in robotic manipulation. MUSIC~\cite{Zhao2021MUSIC} takes the approach of partitioning the state space into the agent's state and the surrounding state, then maximizing the mutual information between them. Furthermore~\cite{Cho2022UnsupervisedRL} investigate combining MUSIC~\cite{Zhao2021MUSIC} with DADS~\cite{Sharma2019DynamicsAwareUD} and multiplicative compositional policy architecture~\cite{MCP} to encourage acquisition of transferable manipulation skills.

While MUSIC-based methods can help agents learn how to interact with objects in their environment, exploration remains challenging. 
Indeed, assuming the agent doesn't interact with changing parts of its surroundings, the quantity of information to learn from is limited.
More recently, controllability-aware skill discovery~\cite{Park2023CSD} proposes a framework improving upon distance-maximizing skill discovery~\cite{Park2022LSD} that encourages actively seeking ``hard-to-achieve'' skills, showing impressive capability to acquire useful robotic manipulation skills. 
In this paper, we approach this problem by augmenting latent-variable skill discovery with additional rewards that improve exploration efficiency, as well as incorporating safety constraints while focusing on the effective combination of multiple rewards with the multi-critic scheme to avoid interference~\cite{Shi2023ReconRC} between various reward components.

\subsection{Multi Critic Learning}

Multi-critic actor learning~\cite{Mysore2022MultiCriticAL} tackles the multi-task reinforcement learning problem by employing multiple critics for each task reward function. 
This approach was shown to minimize possible interference between multiple-task reward signals and allow for stable policy learning in multi-task reinforcement learning.  
Their approach was studied and motivated by the context of multi-style learning in games. Additionally, the usage of multiple critics has been widely studied in various reinforcement learning contexts, such as for tackling overestimation in value-based reinforcement learning ~\cite{DoubleQ, TD3, Lan2020MaxminQC, REDQ}, or for stabilizing learning with uncertainty estimation~\cite{Lee2020SUNRISEAS, Wu2021UncertaintyWA, Lee2021GeneralizableIL}. We differ in our motivation for utilizing multiple critics in this paper, as we are more interested in a multi-objective reinforcement learning viewpoint, particularly in the context of robotic manipulation skill discovery. 
To the best of our knowledge, we are the first to propose utilizing the multi-critic architecture for skill discovery with multiple objectives or constraints in a robotic manipulation framework and demonstrate its effectiveness.

\section{Approach}

\subsection{Preliminaries}

\textbf{Skill Discovery} encompasses unsupervised approaches to reinforcement learning which enable the acquisition of diverse behaviors of a reinforcement learning agent in its environment without specific task rewards. 
One main approach to this problem relies on mutual information maximization between a latent variable sampled from a fixed distribution $p(z)$ which encodes \textit{skills} and states visited by a policy conditioned on this skill~\cite{Gregor2016VariationalIC, Eysenbach2018DiversityIA}. 
This is usually achieved by variational information maximization~\cite{Barber2003TheIA}, by maximizing the following bound:

\begin{align}
    \max \, I(s; z) &= H(z) - H(z|s) \nonumber \\
    &\geq \mathbb{E}_{(s, z)}[\log q_{\eta}(z|s)],
\label{eq:mi}
\end{align}

where $z \sim p(z)$ represents skills, $s$ represents states from an agent's trajectory $\tau = (s_0, ..., s_T)$, and $q_{\eta}(z|s)$ is a discriminator network approximating the posterior distribution of skills given states.

To improve mutual information based skill discovery for learning dynamic skills, Lispchitz-constrained unsupervised skill discovery (LSD)~\cite{Park2022LSD} proposes the maximization of the objective:

\begin{align}
    J_{\text{LSD}} = \mathbb{E}_{\tau, z}\big[\phi(s_{T}) - \phi(s_{0})\big]^{\text{T}}z \nonumber \\ \text{s.t.} \;  \|\phi(s_{T}) - \phi(s_{0})\| \leq  \|s_{T} - s_{0}\|.
\label{eq:lsd}
\end{align}

This objective effectively encourages maximal displacement in a learned state representation space $\phi(s)$, constrained by actual state space displacement with a 1-Lipschitz constant, while ensuring diversity by aligning displacement in representation space with latent skill vectors. In~\cite{Park2022LSD}, $J_{\text{LSD}}$ is decomposed using a telescoping sum $\mathbb{E}_{\tau, z}\big[\sum_{t = 0}^{t = T - 1}\phi(s_{t + 1}) - \phi(s_{t})\big]^{\text{T}}z$ to derive per-transition rewards for a skill-conditioned policy $\pi(a | s, z)$, and the Lipschitz constraint was implemented using Spectral Normalization~\cite{Miyato2018SpectralNF}. $\phi$ and $\pi$ are jointly learned using stochastic gradient descent and reinforcement learning.

\textbf{Multi-critic Actor Learning} is a model-free reinforcement learning approach targeting composite reward function (\emph{i.e.} function with multiple reward components). The approach uses multiple critics, one per reward component, in combination with a single actor in an actor-critic reinforcement learning paradigm introduced in~\cite{Mysore2022MultiCriticAL}. 
Practically, for an actor-critic algorithm using a policy gradient optimization approach, the optimization objective is:

\begin{equation}
    J_{\pi} \propto \mathbb{E}_{\tau, \pi}\big[\text{log} \, \pi(a|s)\sum_{i}\omega_i A_i\big],
\label{eq:mcactor}
\end{equation}

where $A_i$ represents advantage functions for each reward component and $\omega_i$ represents weights used to combine these signals. 
In the seminal multi-critic approach, the authors' primary focus was on learning one critic at a time during updates. 
They achieved this by utilizing sparse encoding for the weights based on the task being learned. 
However, they also conducted preliminary experiments to showcase the effectiveness of using equal weights, demonstrating the ability to interpolate between tasks. 
We build on this approach in our method and adapt it to self-supervised skill discovery.

\subsection{Method}
\label{sec:method}

We consider a Markov Decision Process~\cite{Puterman1994MarkovDP} augmented with a skill latent space  in the domain of robotic manipulation $\mdp{M} = \bigl \langle \mdp{S}, \mdp{A}, \mdp{P}, \mdp{R}, \mdp{Z}\bigr \rangle$, where $\mdp{S}$ is the state space with state vectors $s \in \mathbb{R}^{42}$ containing robot joint positions, robot joint velocities, object pose, end-effector pose, object linear velocity, object angular velocity,  end-effector linear velocity and end-effector angular velocity. We note here that we consider both cartesian positions \textit{and} orientations in object and end-effector poses.
$\mdp{A}$ is the action space of  actions $a \in \mathbb{R}^{7}$ split into two parts: $ a_{\text{arm}} \in [-1, 1]^6$ corresponding to normalised delta pose of the robot's end-effector in Cartesian space which is converted to joint torques using operational space control (OSC)~\cite{Khatib1987AUA}, and ${a_{\text{gripper}}} \in \{0, 1\}$ a Boolean action to open or close the gripper. 
$\mdp{P}$ is the transition function defining our environment dynamics, $\mdp{R}$ is the reward function, and $\mdp{Z}$ is a continuous latent space representing skills.
To enable the discovery of meaningful and safe interaction skills, we define $\mdp{R}$ as a composite reward function consisting of the following reward components: 

\begin{equation}
        r_{\text{reach}} = \frac{1}{\big\| \text{ee\_pos}_t - \text{targ\_pos}_t \big\|_{2}^{2} + \epsilon},
\label{eq:reach}
\end{equation}
where targ\_pos is a pre-specified position of interest, for example, an object's position, ee\_pos is the robot's end effector position, and $\epsilon$ is a threshold for numerical stability,

\begin{equation}
    r_{\text{discovery}} = \big(\phi(s_{t+1}) - \phi(s_{t})\big)^{T}z_t,
\label{eq:discovery}
\end{equation}
where we follow the formulation in LSD~\cite{Park2022LSD} that decomposes the trajectory level reward into per transition rewards using a telescoping sum,

\begin{equation}
r_{\text{safety}} = -\mdp{I}(s_t),
\label{eq:safety}
\end{equation}
where $\mdp{I}$ is a safety indicator function over the states encoding predefined safety constraints which are agent and environment dependent.
In the context of robotic manipulation, such constraints involve joint positions, joint velocities, self-collision avoidance, end-effector velocity, and workspace limits. 
In the experimental section, we detail the necessity of these three components to develop a viable skill-conditioned policy for contact-rich manipulation scenarios.

We propose to use a multi-critic actor learning architecture~\cite{Mysore2022MultiCriticAL} with three critics for the above reward functions to learn a latent variable skill-conditioned policy $\pi(a|s, z)$ using PPO~\cite{schulman2017proximal}.
Fig.~\ref{fig:slim_schematic} illustrates our method and, as far as our knowledge goes, this proposition hasn't been considered in the context of skill discovery. 
Specifically, we propose to utilize a fully separate multi-network architecture as it was shown to perform better in~\cite{Mysore2022MultiCriticAL}. 
One key component in our implementation is that we learn the value functions for each reward function using their respective reward scales but perform a batch normalization of the advantages computed from each critic before combining them with weights for actor learning. 
This scheme has the advantages of (i) fostering unperturbed critic learning per reward component and (ii) easing the burden of choosing appropriate weights to ensure contributions from each reward component are well-balanced when updating the policy. Practically, we use equal weights to combine the normalized advantages.
In addition, we follow the skill composition scheme from ~\cite{kim2023safeskill} by selecting a sequence of skills to execute in each episode which encourages learning safe skill composition. The full algorithm is detailed in Algorithm~\ref{alg:slim}.

\begin{algorithm}[h]
\caption{Skill Learning with Multiple Critics}\label{alg:slim}
\begin{algorithmic}[1]
\Require{Reward functions $r_i$, Critics $V_i$,
Policy $\pi_\theta$,
state representation function $\phi$,
normalization function $\nu$}
\Repeat
\State Sample sequence of skills $(z_1, ..., z_n)$ for rollouts
\State Collect trajectories using $\pi_\theta$ and  $(z_1, ..., z_n)$
\State Update $\phi$ with rollout data to maximize Eq.~(\ref{eq:lsd})
\For{$i \in \{\text{reach}, \text{discovery}, \text{safety}\}$}
\State Update $V_i$: $\min_{\forall (s^t, z^t) \in \tau} \| V_i(s^t, z^t) - \sum_{j = t}^{j = T} r_{i}^j\|$
\State Compute advantage $A_i$ with GAE~\cite{Schulman2015GAE}
\EndFor
\For{$i \in \{\text{reach}, \text{discovery}, \text{safety}\}$}
\State Batch normalize advantages $A_i$: $A_i \leftarrow \nu(A_i)$
\EndFor
\State Update $\pi_\theta$ with PPO, using Eq.~(\ref{eq:mcactor}) with $\omega_i=1$ 
\Until{convergence}
\end{algorithmic}
\end{algorithm}

\section{Experiments}
\label{sec:experiments}
In the context of robotic manipulation, we aim to answer the following questions: 
(Q1) Does SLIM discover \textit{more meaningful} skills than state-of-the-art skill discovery methods? 
(Q2) Does SLIM enable \textit{effective combination of multiple rewards} for skill discovery?
(Q3) Do skills discovered by SLIM lead to \textit{improved learning speed} on downstream tasks?
(Q4) Can skills discovered by SLIM be sequenced to perform \textit{complex} downstream tasks?

To answer our experimental questions, we proceed in four steps.
First, we evaluate sampled rollouts of skill discovery methods to assess their respective diversity in object interaction and safety. 
Second, we evaluate the capabilities to train safe and efficient motor primitives with Hierarchical Reinforcement Learning (HRL)~\cite{Barto2003RecentHRL}.
In detail, we evaluate our approach on position and orientation matching, which implicitly involves behaviors like reaching, grasping, pushing, and displacing.
Third, we leverage our HRL-trained motor primitives with a planner to validate our approach for safe object-centric trajectory tracking. 
Finally, we extend our trajectory tracking evaluation for multiple object manipulation.

\textbf{Setup}
We use a tabletop manipulation environment modeled in the IsaacGym simulator~\cite{Makoviychuk2021IsaacGH} illustrated in Fig.~\ref{fig:slim_sim_topview} and Fig.~\ref{fig:slim_sim_sideview}. The environment includes a Franka Emika Panda robot, a table, and a 5-cm cube. The robot is mounted on the table and is always initialized in a fixed configuration shown in the side view image in Fig.~\ref{fig:slim_sim_sideview}. Meanwhile, the object is initialized at a randomly sampled position within an initialization area  of dimension 24 x 24 cm, illustrated with the orange square in Fig.~\ref{fig:slim_sim_topview}. The object's orientation is also initialized randomly using a uniform distribution over axis angle rotations. Compared to tabletop manipulation setups studied in previous works~\cite{Zhao2021MUSIC, Cho2022UnsupervisedRL, Park2022LSD, Park2023CSD}, our setup is more challenging as our initial object poses are sampled from a wider distribution and our action space is larger\footnote{These works usually consider the Fetch robotics manipulation environments~\cite{Plappert2018MultiGoalRL} where the gripper orientation is fixed and object initial orientations are also fixed to be aligned with the gripper. As such the actions only control 3-D cartesian displacements while we control 6-D position and orientation displacements.}. In our experiments, we leverage the high level of parallelism enabled by IsaacGym by running 5000 instances of our environment in parallel. Furthermore, for all experiments, we use a 6-D von Mises-Fisher distribution as the fixed prior skill distribution. Intuitively, these skills correspond to representing position and orientation displacements.

\textbf{Baselines}
We use the following skill discovery methods as baselines: DIAYN~\cite{Eysenbach2018DiversityIA}, LSD~\cite{Park2022LSD} and SASD~\cite{kim2023safeskill}. 
DIAYN and LSD are chosen to serve as commonly used and cited latent variable skill discovery methods. 
SASD serves as a baseline that introduces the safe skill discovery formalism and tackles both objectives of skill discovery and safety.

\begin{figure}[ht!]
\begin{subfigure}{\textwidth}
    \begin{subfigure}{0.15\textwidth}
    \centering
    \includegraphics[width=\linewidth]{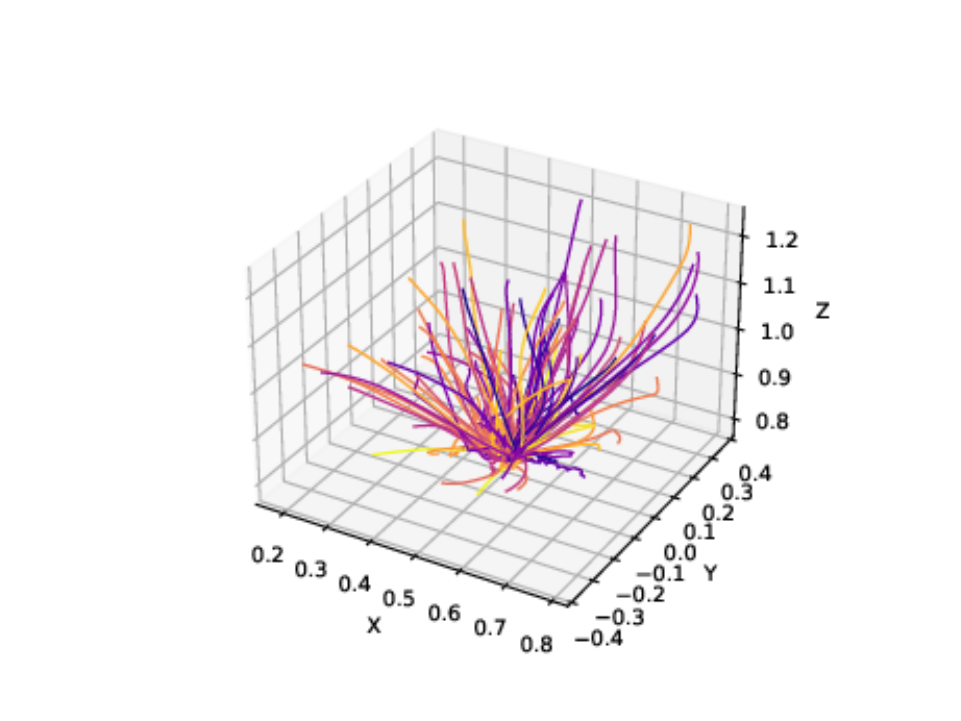}
    \caption{SLIM}
    \label{subfig:slim_traj1}
    \end{subfigure}
    \begin{subfigure}{0.15\textwidth}
    \includegraphics[width=\linewidth]{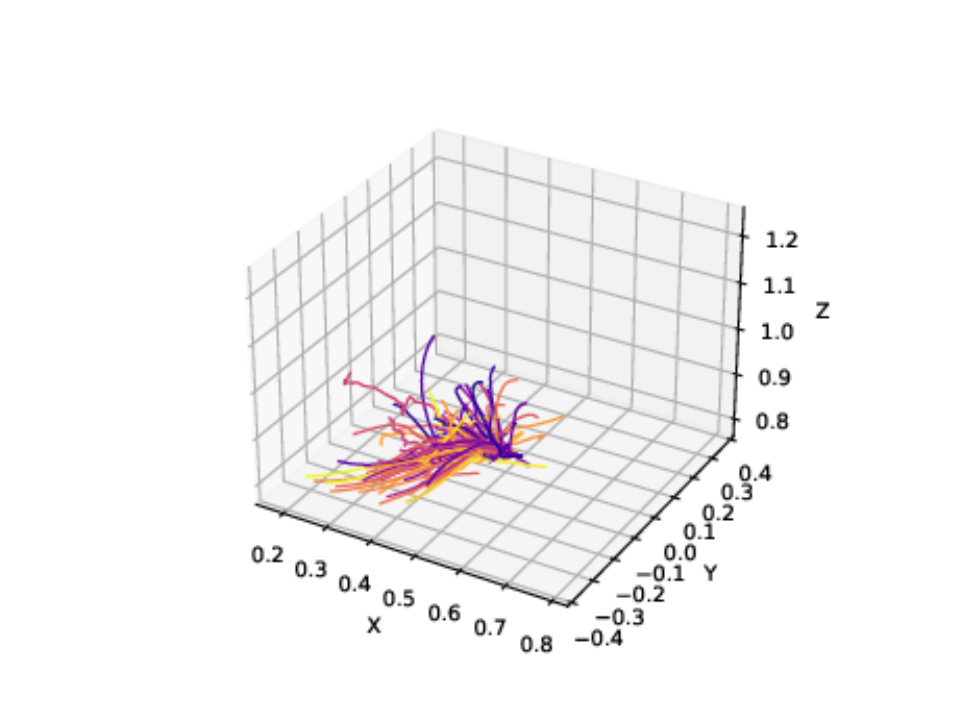}
    \caption{SLIM\_ur}
    \label{fig:multi_reward_unnormalized}
    \end{subfigure}
    \begin{subfigure}{0.15\textwidth}
    \includegraphics[width=\linewidth]{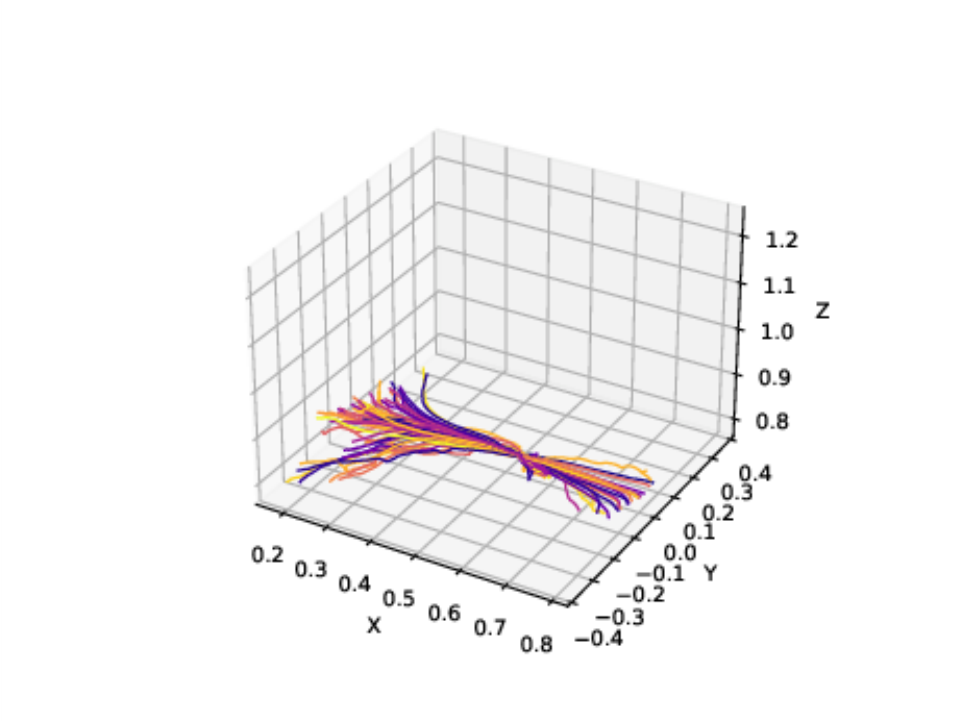}
    \caption{SLIM\_nr}
    \label{fig:multi_reward_normalized}
    \end{subfigure}
\end{subfigure}

\begin{subfigure}{\textwidth}
    
    \begin{subfigure}{0.15\textwidth}
    \includegraphics[width=\linewidth]{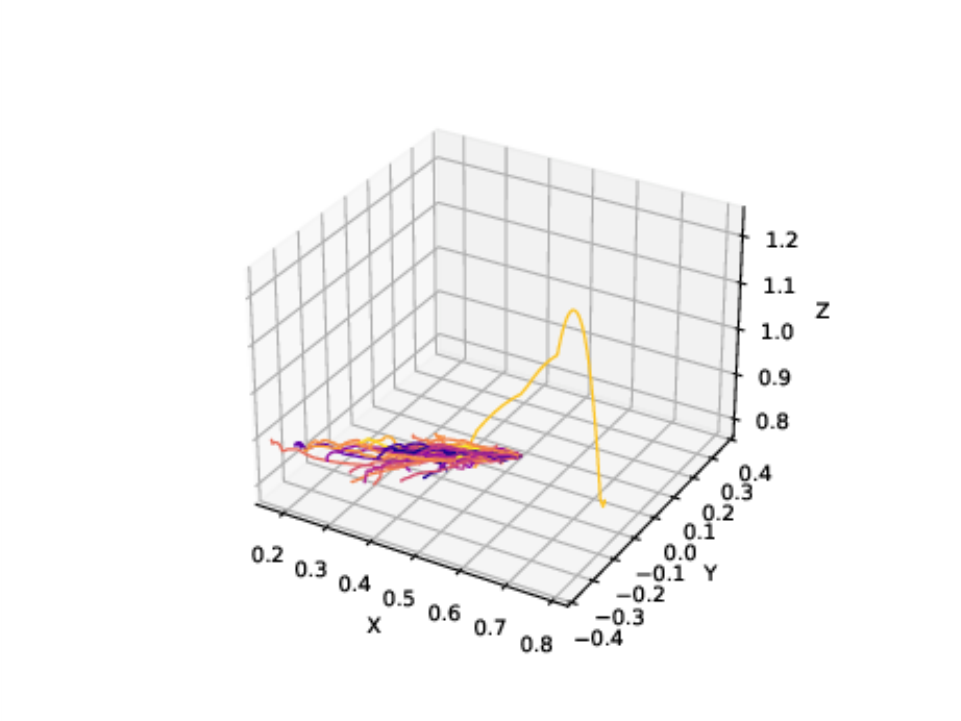}
    \caption{SLIM\_no\_r}
    \label{subfig:slim_noreach_traj}
    \end{subfigure}
    \begin{subfigure}{0.15\textwidth}
    \includegraphics[width=\linewidth]{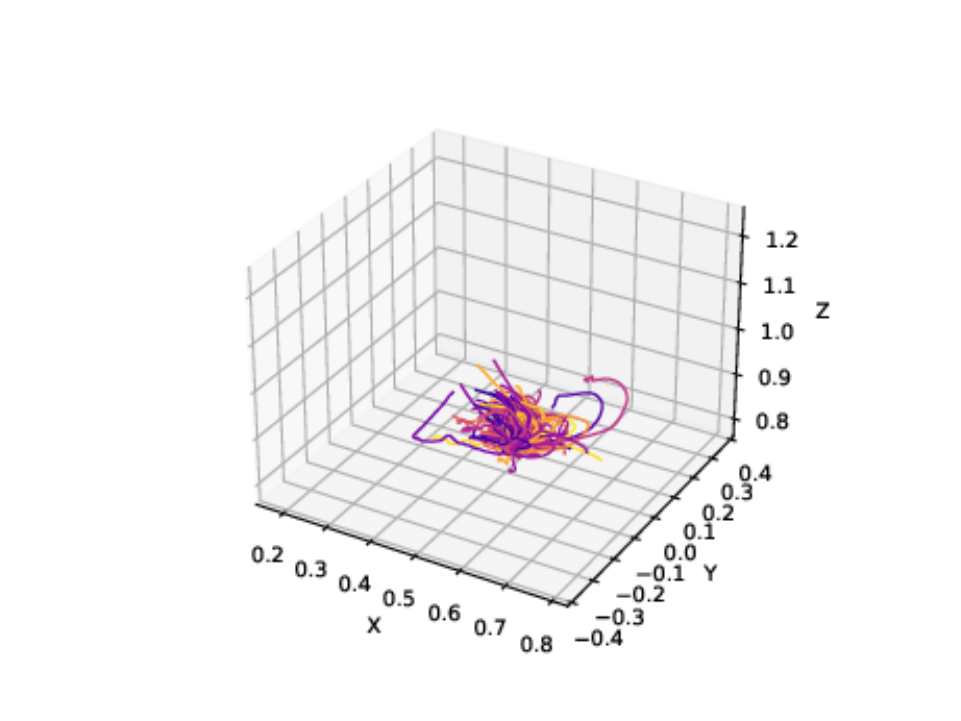}
    \caption{SLIM\_no\_d}
    \label{subfig:slim_nodiscovery_traj}
    \end{subfigure}
    \begin{subfigure}{0.15\textwidth}
    \centering
    \includegraphics[width=\linewidth]{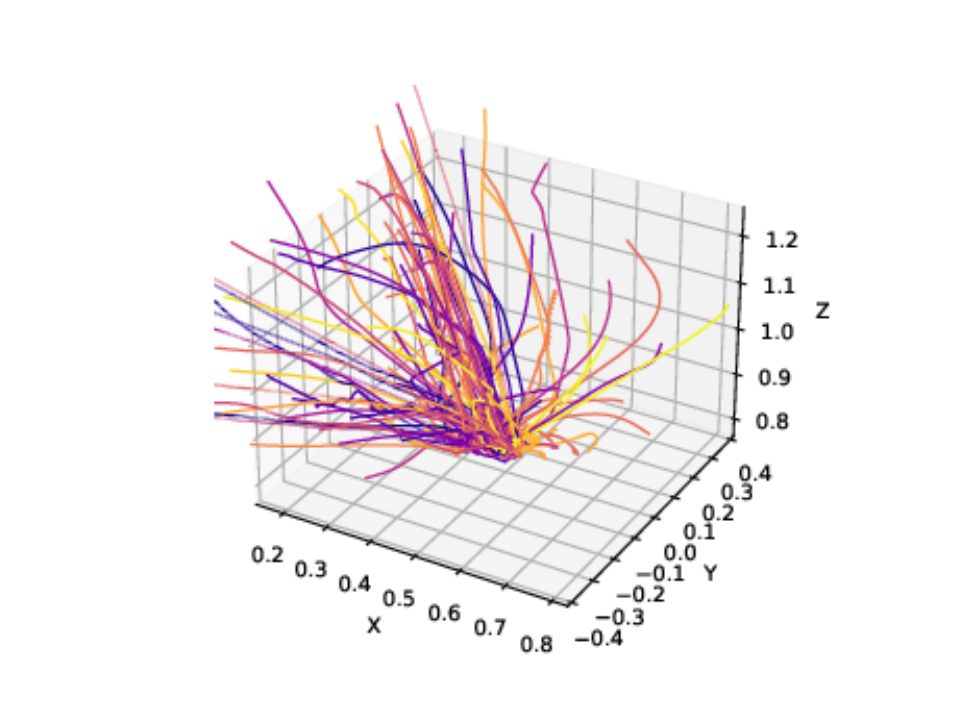}
    \caption{SLIM\_no\_s}
    \label{subfig:slim_nosafety_traj}
    \end{subfigure}
\end{subfigure}

\begin{subfigure}{\textwidth}
    \begin{subfigure}{0.15\textwidth}
    \centering
    \includegraphics[width=\linewidth]{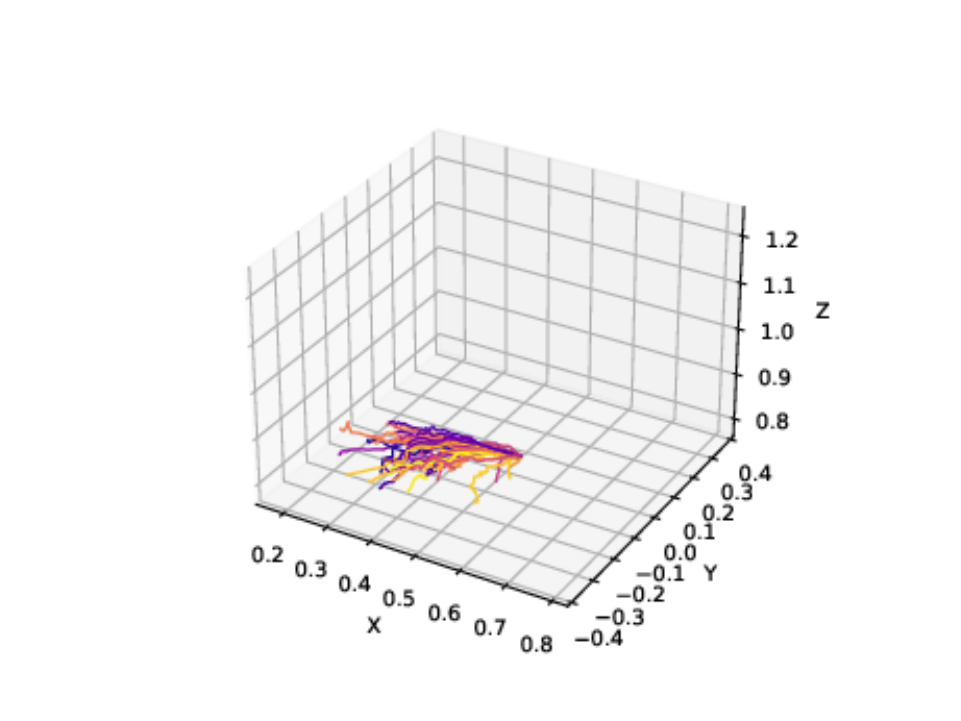}
    \caption{SASD}
    \label{subfig:sasd}
  \end{subfigure}
  \begin{subfigure}{0.15\textwidth}
    \centering
    \includegraphics[width=\linewidth]{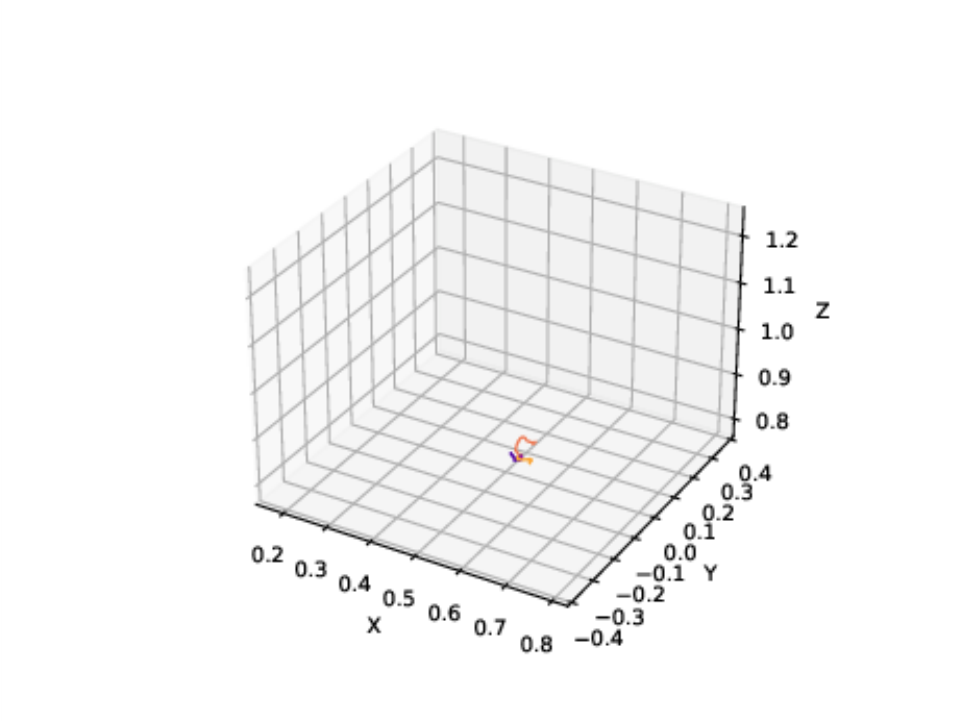}
    \caption{DIAYN}
    \label{subfig:diayn}
  \end{subfigure}
  \begin{subfigure}{0.15\textwidth}
    \centering
    \includegraphics[width=\linewidth]{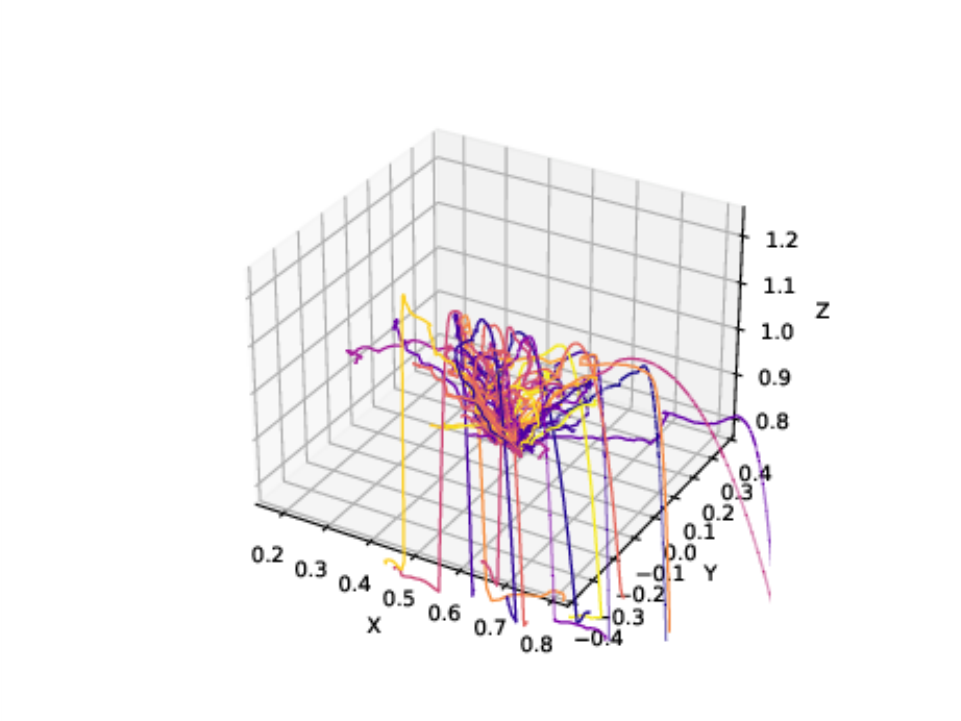}
    \caption{LSD}
    \label{subfig:lsd}
  \end{subfigure}
\end{subfigure}
  
    \caption{\textbf{Skill trajectories for SLIM, SLIM ablations, and baselines}. SLIM outperforms baselines in terms of grasping consistency and the diversity of the cube's displacement. The baselines do not learn to pick up the cube. While SLIM ablations show different levels of object interaction with both picking and pushing behaviors emerging, only SLIM learns interactive, diverse and safe displacement manipulations}
    \label{fig:sd_baselines}
\end{figure}

\subsection{Qualitative evaluation}
 For our qualitative evaluation in Fig.~\ref{fig:sd_baselines} we plot color-coded 3-D object trajectories over $200$ environment steps with the skill-conditioned polcies for $100$ randomly sampled skill vectors $z$ .
 
\textbf{SLIM vs. Baselines}: From Fig.~\ref{fig:sd_baselines} we observe that LSD learns to push the object but quite unsafely as the object gets knocked off the table frequently. 
On the other hand, SASD learns safe pushing behaviors but fails to grasp and lift, while DIAYN hardly interacts with the object.
SLIM outperforms both baselines as even though they both learn some form of pushing, they all fail to learn grasping and lifting in many directions.

\textbf{SLIM vs. SLIM ablations}: To better understand the effectiveness of the proposed method, we compare SLIM to various ablated versions. These ablations can be grouped in two groups. The first group considers using the same reward functions as SLIM, Eq.~(\ref{eq:reach}), Eq.~(\ref{eq:discovery}), Eq.~(\ref{eq:safety}), but combined into a single reward function (by simple summation) and hence a single critic. Note that this provides all the same reward signals used in SLIM except we perform the weighted combination of the rewards before learning a single critic. 
Our first ablation, henceforth called \textbf{SLIM\_unnormalized\_rewards} (a.k.a SLIM\_ur) consists of summing up all rewards. 
In Fig.~\ref{fig:multi_reward_unnormalized}, the skill policy rollouts with this method shows some grasping and lifting behavior is learned but the trajectories are less diverse than in SLIM which uses multiple critics.
Furthermore, to prevent different reward scales to be determinants of performance differences, we define a second ablation version called \textbf{SLIM\_normalized\_rewards} (a.k.a SLIM\_nr).
Here, similar to SLIM, we apply normalization to ensure all reward signals are on similar scales before combining them into a single reward function. 
The trajectories from this version are visualized in Fig.~\ref{fig:multi_reward_normalized} showing less diversity than SLIM and mostly failing to learn grasping and picking.

The second ablation group considers subset combinations of the reward functions namely: \textbf{SLIM\_no\_reach} (a.k.a. SLIM\_no\_r): using $r_{\text{discovery}}$ and $r_{\text{safety}}$, \textbf{SLIM\_no\_discovery} (a.k.a SLIM\_no\_d): using $r_{\text{reach}}$ and $r_{\text{safety}}$, and \textbf{SLIM\_no\_safety} (a.k.a SLIM\_no\_s): using $r_{\text{reach}}$ and $r_{\text{discovery}}$. \textbf{SLIM\_no\_reach} in Fig.~\ref{subfig:slim_noreach_traj} shows the effect of combining safety with LSD is very similar to SASD. We observe the safety reward constrains LSD from knocking objects off the table but with limited diversity of object displacements. On the other hand, \textbf{SLIM\_no\_discovery} in Fig.~\ref{subfig:slim_nodiscovery_traj}  shows safe object manipulations with some grasping, pushing and lifting, but quite limited diversity due to the missing discovery reward component. Finally, for \textbf{SLIM\_no\_safety} in Fig.~\ref{subfig:slim_nosafety_traj}, we observe that the robot learns to displace the object in multiple directions showing a very effective combination of reaching and distance maximization discovery rewards similar to SLIM, but is unconstrained by the safety component hence it learns to over-extend the robot in order to maximize the discovery component leading to unsafe robot configurations.

Overall, our ablations show the importance of each component to obtain diverse yet interactive and safe manipulation behaviors. We observe that the three reward components are necessary and complementary to achieve our desired behaviors. In the first group of ablations, we clearly observe the difficulty with combining these three components using normalized or unnormalized sums due to possible interferences between reward signal while learning a skill-conditioned policy. Utilizing the multi-critic architecture with dedicated criticis per reward component helps to alleviate this problem and stabilize learning. Furthermore, we show with the second ablation group that while the multi-critic scheme helps with combining reward components, an omission of any of the three rewards, Eq.~(\ref{eq:reach}), Eq.~(\ref{eq:discovery}), Eq.~(\ref{eq:safety}), hampers the overall result.

\subsection{Quantitative evaluation}

\begin{figure}[h!]
\begin{subfigure}{0.4\textwidth}
    \centering
    \includegraphics[width=1.0\linewidth]{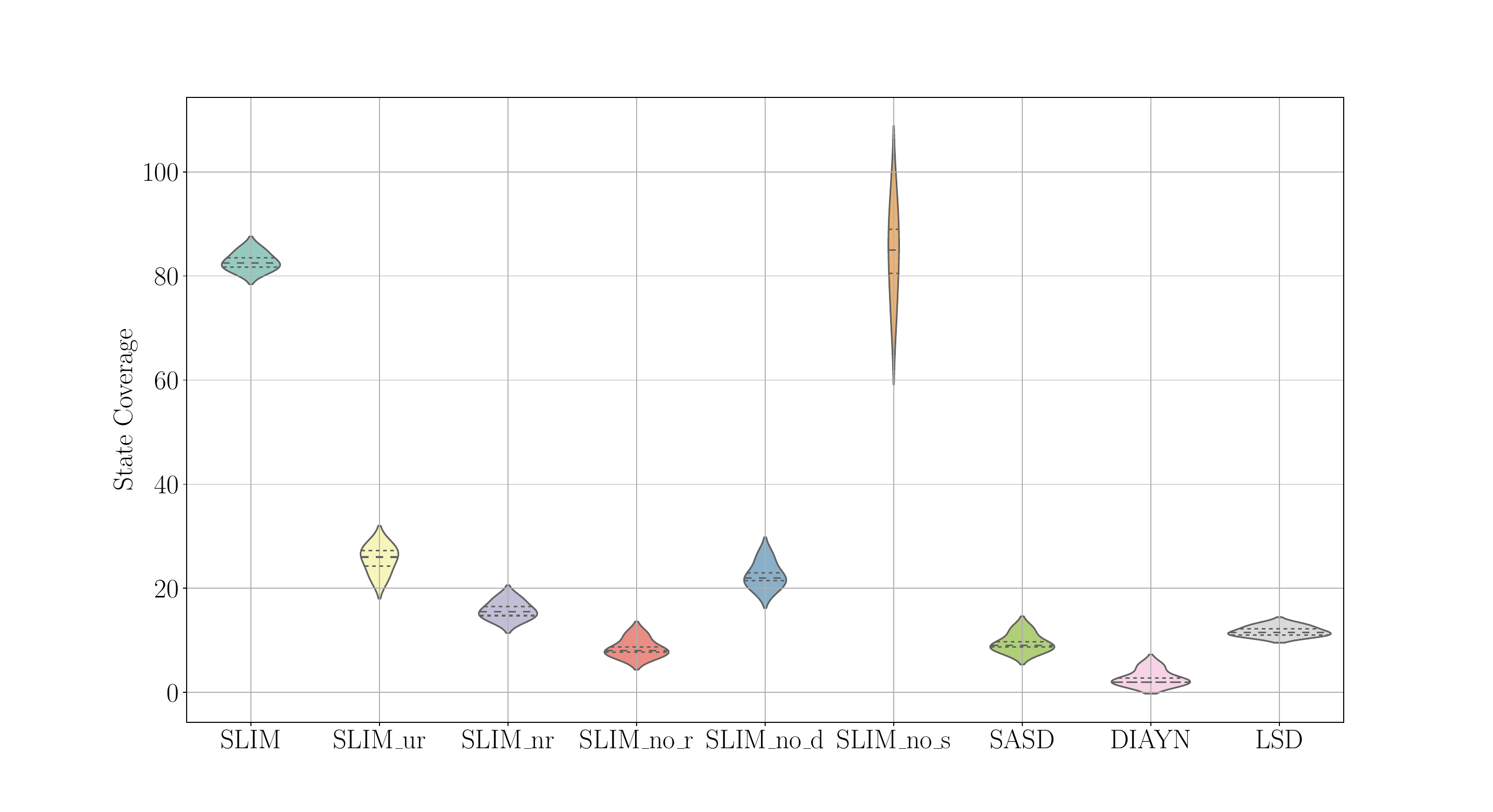}
    \caption{Coverage}
    \label{fig:skills_coverage}
\end{subfigure}
\begin{subfigure}{0.4\textwidth}
    \centering
    \includegraphics[width=1.0\linewidth]{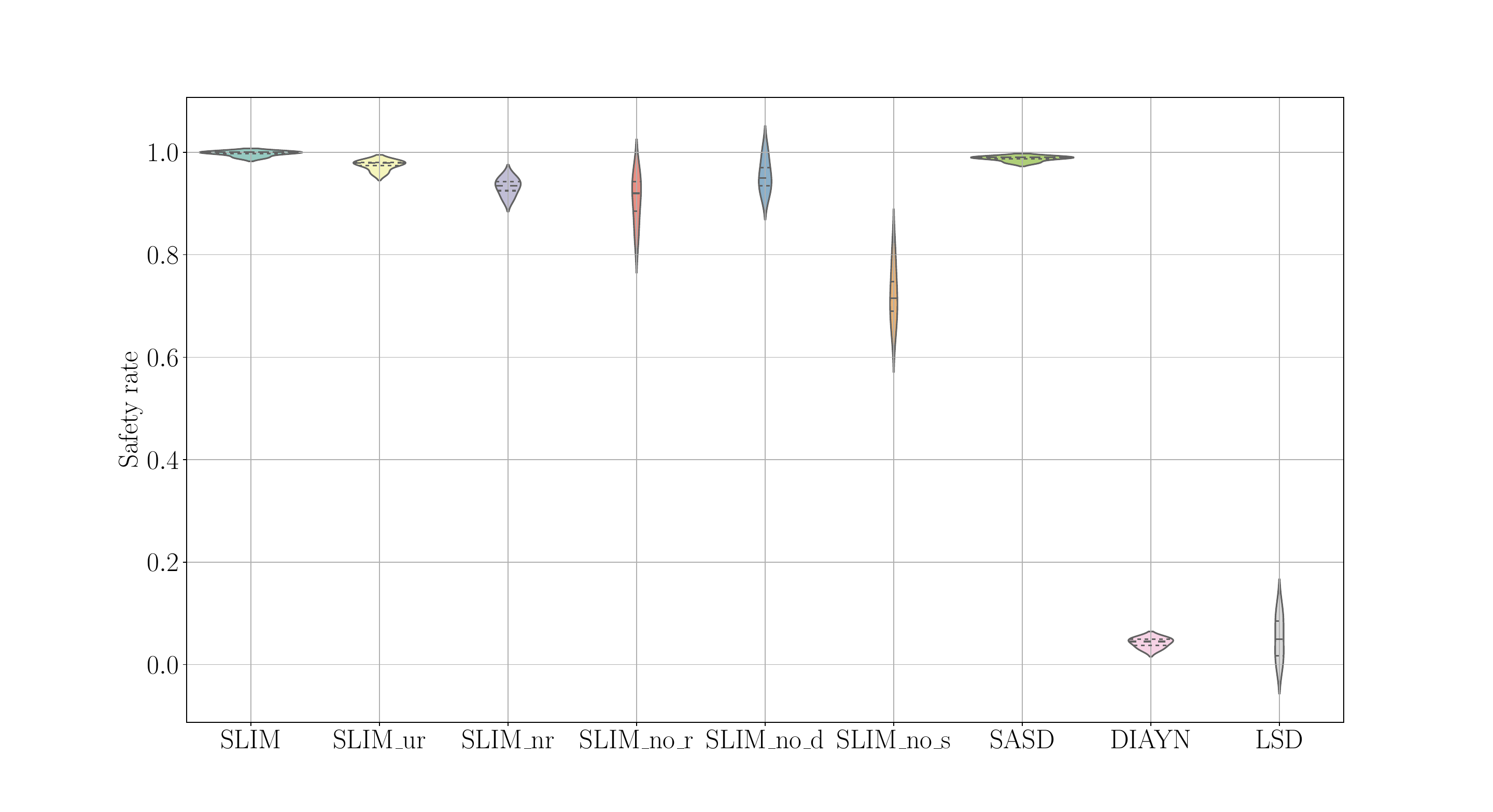}
    \caption{Safety}
    \label{fig:skills_safety}
\end{subfigure}
    
\caption{\textbf{Coverage and Safety}. Coverage is the number of boxes discretizing the workspace covered by the object. Safety is the ratio of safe states encountered during random skill rollouts.}
    \label{fig:coverage}
\end{figure}

\textbf{Coverage and Safety} We evaluate coverage and safety for SLIM, SLIM ablations, and baselines. Coverage is measured over a 50 x 50 x 50 cm centered region discretized into 125 units of 10cm cubes. We evaluate by rolling out 100 trajectories per method repeated for 4 seeds and visualize the mean and standard deviation of the number of cubes covered by the object during the rollouts. Safety is measured as the ratio of safe states according to the indicator function in~Eq.~(\ref{eq:safety}) encountered during the rollouts. Both measures are visualized as violin plots shown in Fig.~\ref{fig:coverage}. We observe that SLIM matches the strongest safety baseline SASD while being significantly superior in coverage to baselines.

\begin{figure}[h!]
    \begin{subfigure}{0.4\textwidth}
    \includegraphics[ width=1.0\linewidth]{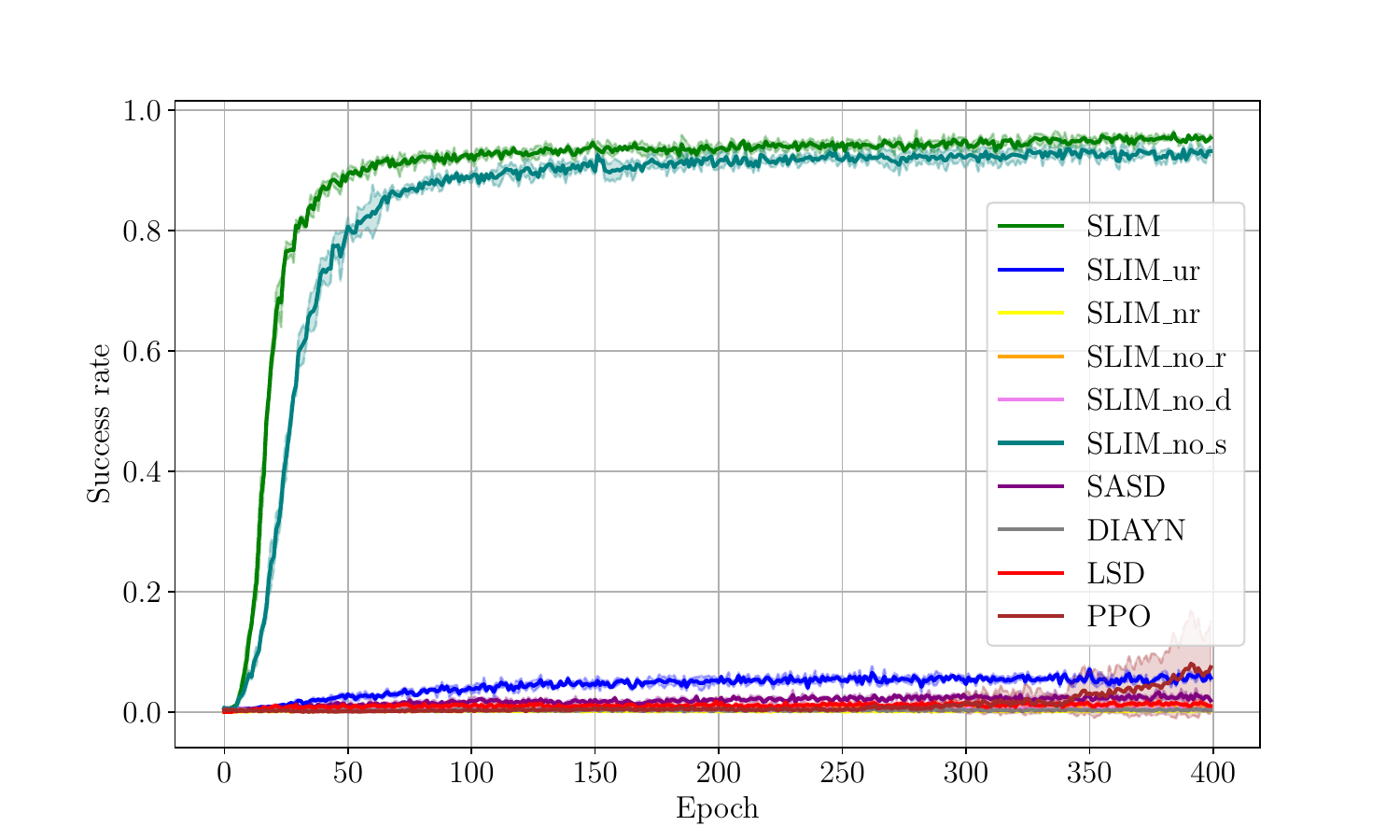}
    \caption{Position matching}
    \label{fig:position_matching}
    \end{subfigure}
    \begin{subfigure}{0.4\textwidth}
    \includegraphics[ width=1.0\linewidth]{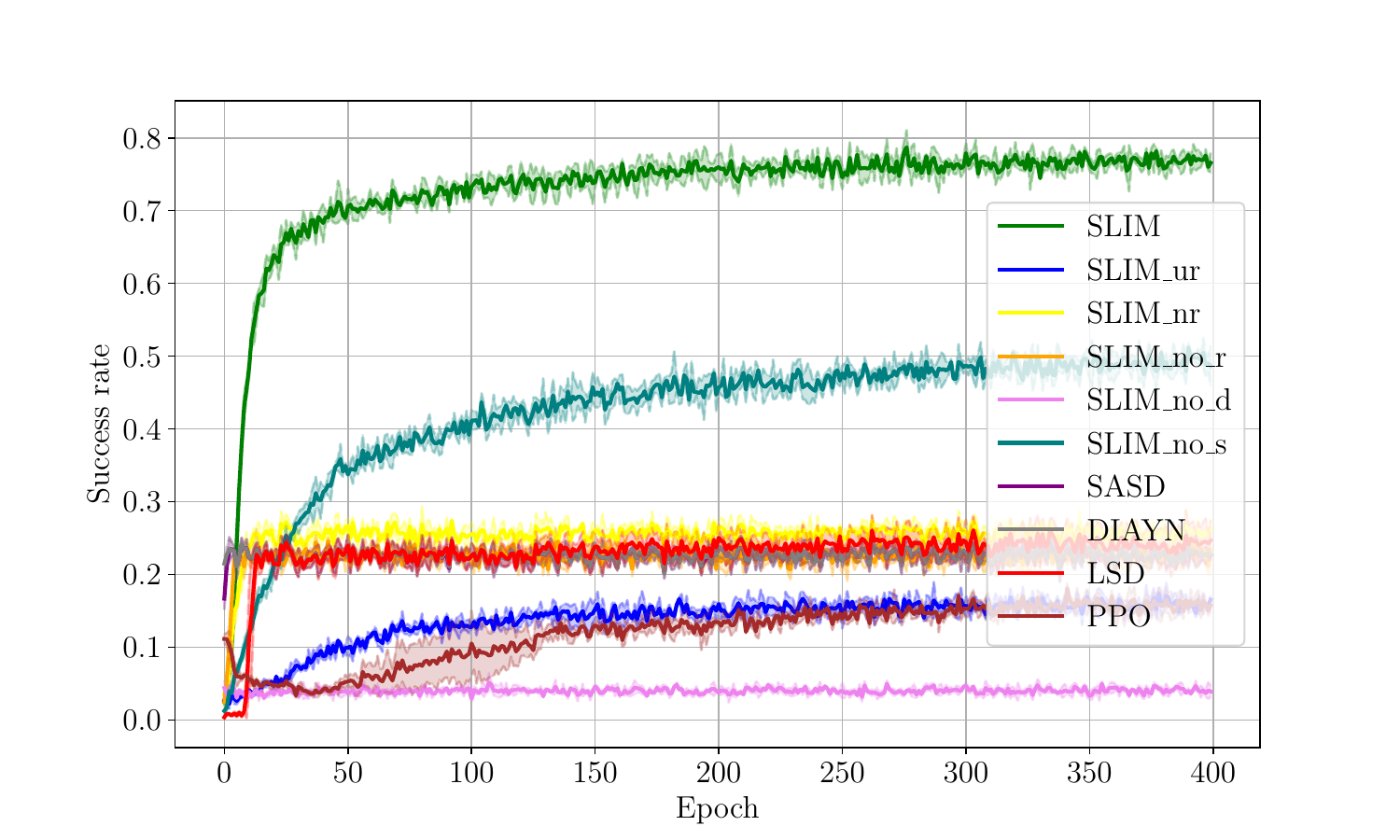}
    \caption{Orientation matching}
    \label{fig:orientation_matching}
    \end{subfigure}
    \caption{\textbf{Performance on downstream tasks}. We evaluate our approach with the position-matching and orientation-matching tasks. SLIM enables improved sample efficiency across all downstream tasks}
    \label{fig:downstream}
\end{figure}

\textbf{Skill utility for downstream tasks} To assess the utility of discovered skills across all skill discovery methods introduced above, we train a hierarchical controller with HRL above the skill-conditioned policies to solve downstream robotic manipulation tasks. Specifically, we evaluate our approach to the tasks of position-matching and orientation-matching. We chose these two tasks because they correspond to the prime competencies required in robotic manipulation for re-arrangement type tasks. Furthermore, they illustrate how well the full range of skills learned over object position and orientation displacements can be leveraged. We compare SLIM to our baselines, SLIM ablations, and reinforcement learning from scratch with PPO. From Fig.~\ref{fig:downstream} we observe that only skills learned by SLIM (and SLIM\_no\_safety) can be leveraged by the hierarchical controller to solve both tasks with vastly improved sample efficiency.

\begin{figure}
    
    \begin{subfigure}{\textwidth}
    % \centering
    \begin{subfigure}{0.15\textwidth}
    % \centering
    \includegraphics[height=2.5cm, width=1\linewidth]{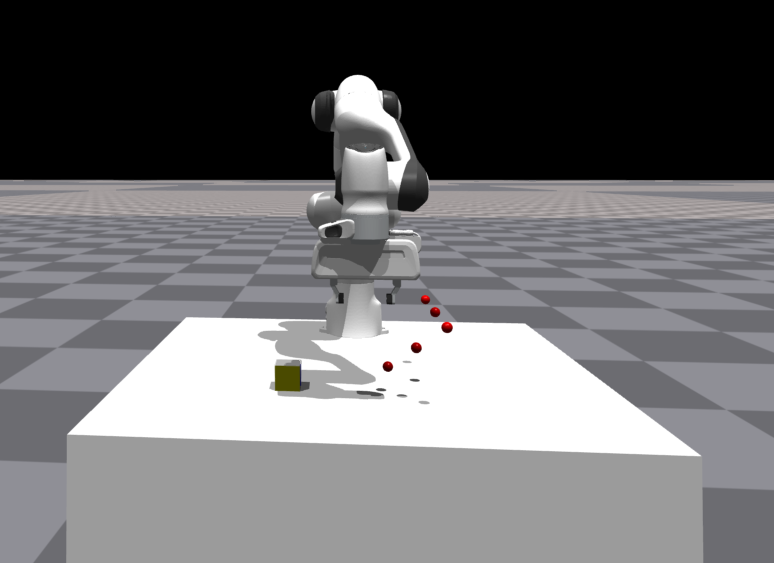}
    \label{fig:traj1}
    \end{subfigure}
    \begin{subfigure}{0.15\textwidth}
    % \centering
    \includegraphics[height=2.5cm, width=1\linewidth]{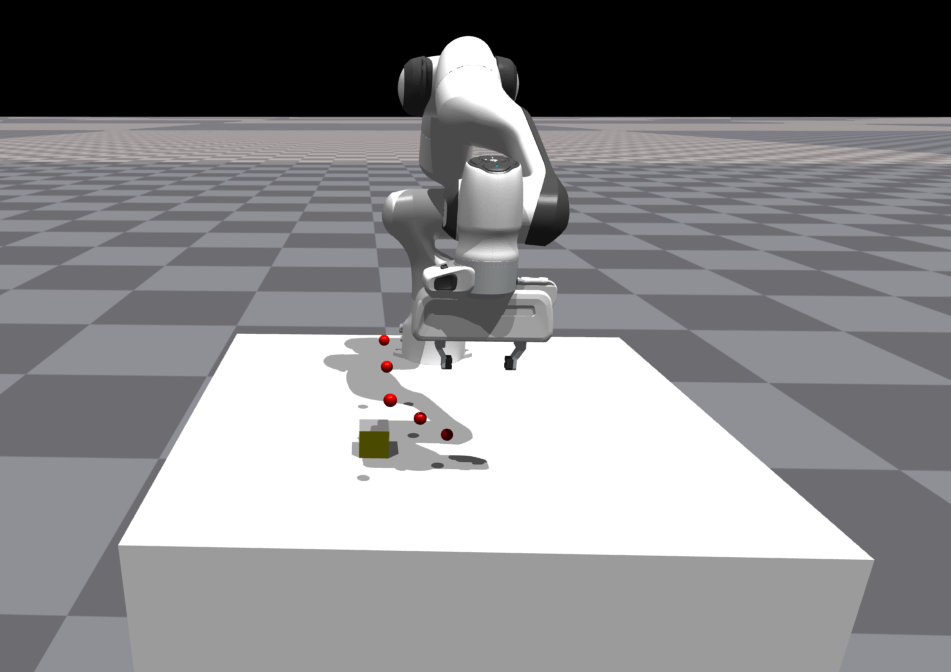}
    \label{fig:traj2}
    \end{subfigure}
    \begin{subfigure}{0.15\textwidth}
    % \centering
    \includegraphics[height=2.5cm, width=1\linewidth]{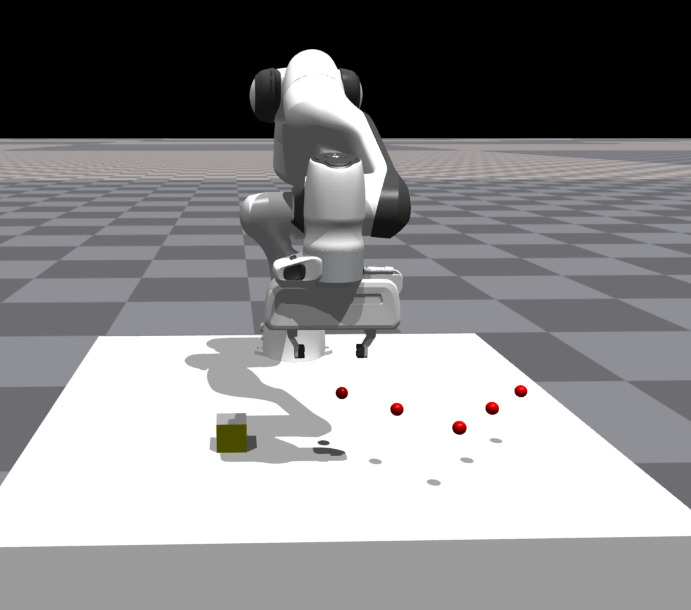}
    \label{fig:traj3}
    \end{subfigure}
    \end{subfigure}
    \begin{subfigure}{\textwidth}
    % \centering
    \begin{subfigure}{0.15\textwidth}
    % \centering
    \includegraphics[height=2.5cm, width=1\linewidth]{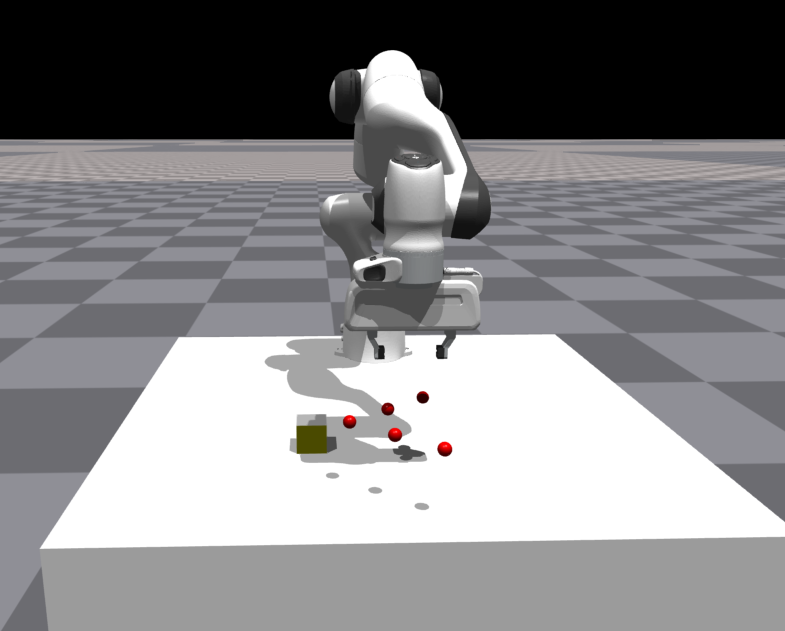}
    \label{fig:traj4}
    \end{subfigure}
    \begin{subfigure}{0.15\textwidth}
    % \centering
    \includegraphics[height=2.5cm, width=1\linewidth]{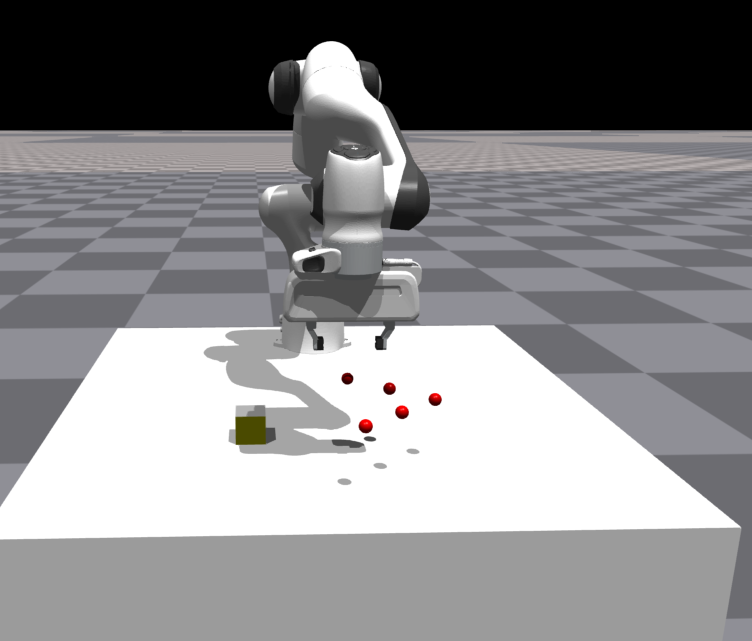}
    \label{fig:traj5}
    \end{subfigure}
    \begin{subfigure}{0.15\textwidth}
    % \centering
    \includegraphics[height=2.5cm, width=1\linewidth]{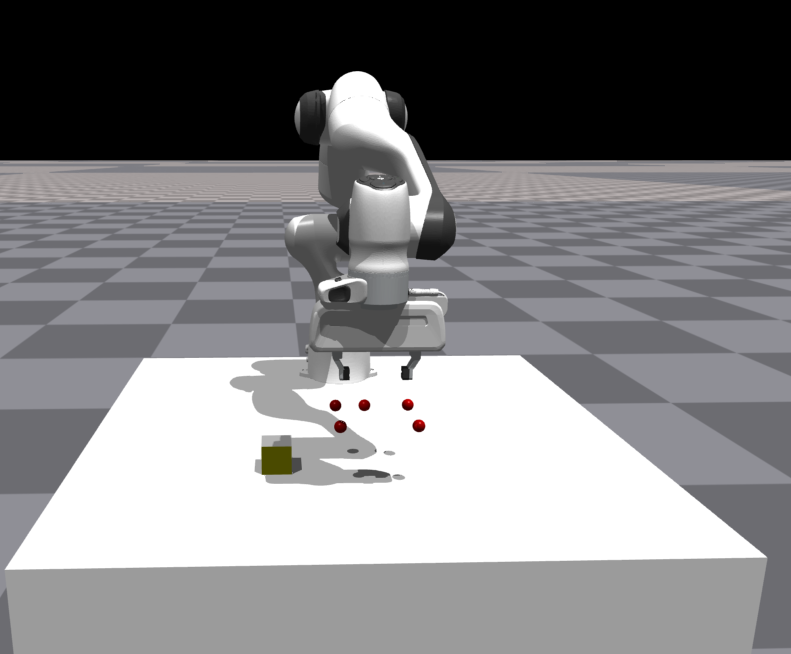}
    \label{fig:traj6}
    \end{subfigure}
    \end{subfigure}
    
    \caption{\textbf{Safe trajectory following}. We evaluate our HRL policies trained over SLIM as motor primitives for safe trajectory following. The six trajectories evaluated are shown in order from top left to bottom right}
    \label{fig:trajectory_following}
\end{figure}

\textbf{Safe object-trajectory following} Next, we investigate the ability to use SLIM for safe object-trajectory tracking, which offers more usability than HRL alone for solving downstream tasks. 
We demonstrate how our skill-based HRL policy, when used as a motor primitive, can be useful. 
Additionally, we examine the impact of errors in this context across six different types of trajectories.
As shown in Fig.~\ref{fig:trajectory_following}, all trajectories are described using five ordered points defined in Cartesian space. 
We roll out the position-matching HRL policy trained in the previous section to follow a trajectory by sequentially selecting the points in order as position-matching goals for the policy. 
We evaluate using the following metrics: 
(i) \textit{Overall success} (\%) indicates if the trajectory is followed successfully by reaching all the points above a given distance threshold of 5cm, 
(ii) \textit{Maximum distance} (m) indicates the maximum distance between the object and the current waypoint at all phases in the trajectory,
(iii) \textit{Points success} indicates the total number of points successfully approached in the trajectory, and
(iv) \textit{Safety rate} (\%) indicates the ratio of safe states encounter over the trajectory.

\begin{table}[h]
\label{table:1}
\caption{Safe object-trajectory following and multi-object rearrangement using SLIM-based motor primitives}
\setlength{\tabcolsep}{2pt}
  \centering
    \begin{tabular}{cccccc}

      \hline
      Trajectory & Overall success  & Max distance & Points success & Safety rate \\
      \hline
      1 & 100 & 0.04 $\pm$ 0.00  & 5 $\pm$ 0.00  & 100 \\
      2 & 100 & 0.04 $\pm$ 0.00  & 5 $\pm$ 0.00    & 100 \\
      3 & 80 & 0.05 $\pm$ 0.03  & 4.5 $\pm$ 1.20    & 99.97 \\
      4 & 60 & 0.07 $\pm$ 0.04 & 4.4 $\pm$ 0.91  & 100 \\
      5 & 80 & 0.12 $\pm$ 0.20  & 4.2 $\pm$ 1.66       & 100 \\
      6 & 80 & 0.05 $\pm$ 0.02  & 4.7 $\pm$ 0.64  & 100 \\
      \hline
      \hline
      Line & 100 & 0.034 $\pm$ 0.009  &  3.0 $\pm$ 0.00  & 100 \\
      Pyramid & 80 & 0.048 $\pm$ 0.014  & 2.8 $\pm$ 0.60    & 99.98 \\
      \hline
      \hline
    \end{tabular}
\end{table}

Our findings, as displayed in Table I, suggest that SLIM-based motor primitives can serve within planning algorithms, offering an inherent level of safety.
This opens the door to executing complex trajectories over single or multiple objects while ensuring arbitrary safety criteria.

\textbf{Multi-object manipulation} Finally, we take our trajectory-following task one step further by evaluating the ability to solve complex downstream tasks involving multiple objects. 
Specifically, we evaluate the same planning-based trajectory following approach but to re-arrange a set of three cubes into various configurations namely: (a) Line: where we align the cubes to the horizontal axis, and (b) Pyramid: where we form a base with two cubes and place the third cube over this base. 
For each cube, we plan a trajectory to reach the end pose in the desired configuration and sequentially execute the trajectory following. 
We evaluate using the same metrics introduced above and the results are also shown in Table I. 
Points success for this case refers to the number of cubes correctly placed in their final pose for the desired configuration.

\section{Conclusion}
\label{sec:conclusion}
In this paper, we have introduced SLIM, a novel approach to skill discovery tailored to the challenges of robotic manipulation.
We empirically demonstrated that by integrating multiple critics and associated reward functions, the resulting skill-conditioned policy acquires safe and diverse manipulation skills that can be leveraged for downstream tasks using hierarchical reinforcement learning and planning.
One limitation of our approach is that we assume an easy-to-design and reasonably generic bonus reward function to help with encouraging object interaction. 
A natural extension for future work is to replace this bonus with another intrinsic reward function that serves the same purpose of easing exploration.
Likewise, we plan to further study the interference between multiple rewards, which necessitates such an approach. 
Additionally, exploring improved compositions of the advantages used in the policy gradient would be an interesting avenue for investigation. 
Lastly, sim2real deployment of our learned skill policies and assessing the benefits of applying our approach in other fields such as locomotion and navigation holds potential for fruitful exploration.

\section*{ACKNOWLEDGMENT}
We thank Taeyoon Lee, Younghyo Park, and the Control and Dynamics team at NAVER LABS for their SASD codebase upon which this research project was developed. We also thank Paul Jansonnie and all team members of Robot Learning at NAVER LABS Europe for insightful discussions and suggestions.

\bibliographystyle{IEEEtran}
\bibliography{references}  % .bib

\end{document}